# Geo-Information Harvesting from Social Media Data

Xiao Xiang Zhu, Yuanyuan Wang, Mrinalini Kochupillai, Martin Werner, Matthias Häberle, Eike Jens Hoffmann, Hannes Taubenböck, Devis Tuia, Alex Levering, Nathan Jacobs, Anna Kruspe, Karam Abdulahhad

*Abstract*—**This is the pre-acceptance version, to read the final version please go to IEEE Geoscience and Remote Sensing Magazine on IEEE XPlore.**
**As unconventional sources of geo-information, massive imagery and text messages from open platforms and social media form a temporally quasi-seamless, spatially multi-perspective stream, but with unknown and diverse quality. Due to its complementarity to remote sensing data, geo-information from these sources offers promising perspectives, but harvesting is not trivial due to its data characteristics. In this article, we address key aspects in the field, including data availability, analysis-ready data preparation and data management, geo-information extraction from social media text messages and images, and the fusion of social media and remote sensing data. We then showcase some exemplary geographic applications. In addition, we present the first extensive discussion of ethical considerations of social media data in the context of geo-information harvesting and geographic applications. With this effort, we wish to stimulate curiosity and lay the groundwork for researchers who intend to explore social media data for geo-applications. We encourage the community to join forces by sharing their code and data.**

## I. Introduction

Geodetically accurate remote sensing (RS) data acquired by Earth observation (EO) satellites serves as a high quality reference database for global geo-information retrieval. Beyond the temporal resolution of EO satellites, typically days, the contextual embedding of space into meanings, perceptions, and dynamic changes in human settlement due to daily life routines can only be indirectly assessed by ground-level measurement, such as social media data. Taking building function prediction as an example, building façades and detailed building functional information retrievable from ground-level social media imagery are not accessible from satellites. Such information can also be utilized to generate training sets for supervised classification with satellite images.

A new era of Earth observation (EO) has certainly arrived, when we consider the social media data (photos, text messages) uploaded by individuals as a valuable additional information source of Earth "observation." As of this writing (April, 2022), around 3.96 billion people use social networking

X. Zhu is with the Chair of Data Science in Earth Observation, Technical University of Munich (TUM), Germany. Email: xiaoxiang.zhu@tum.de (corresponding author). YW, MK, MW, MH, EJH, AK are with TUM, Germany; HT and KA are with German Aerospace Center, Germany; DT is with Ecole Polytechnique Fed erale de Lausanne, Sion, Switzerland; AL is with Wageningen University, the Netherlands; NJ is with Washington University in St. Louis, USA.

sites [1], such as Facebook. As shown in Fig. 1, a subset of selected social media platforms already provided 3 billion daily photo-uploads in 2015 and estimates suggest hundreds of petabytes of them are available in total. Apart from internet images and text messages, 2D geographic information systems (GISs), digital cadastral databases, and municipal utility information are widely available for most well developed countries. Besides paid services, openly available sources of 2D GIS data include Natural Earth, Geocommons, MapCruzin, OpenStreetMap (OSM), and many more. Exploiting and extracting the valuable information from these data sources enables a revolutionary complement to satellite remote sensing. The extracted geo-information from these observations will support cartographic applications, civil security, and city planning, among many other domains, and hence change the way we manage our cities.

Research fields in social media data mining and outsourcing sensing tasks to the general public are rapidly emerging, especially for 3D urban reconstruction from social media imagery [2]–[5], people dynamics monitoring using airborne sensor and mobile phone data [6], [7], flood damage mapping using governmental and crowdsource data [8], [9], and crowdsourcing for mapping, image analysis, and geographic information collection [10]–[13] (and the list goes on). The exponential increase of social media data ignites a new means of remote sensing that involves the community, also known as community remote sensing. The real strength of social media in remote sensing is its complementarity in data characteristics and population base.

Only a handful of contributions have addressed the problem of fusing social media and RS data for geo-information retrieval. As [14] mentions, few studies on social media text messages are linked to remote sensing data. Most of them focus on result-level spatial merging. The current research has not addressed the real challenge of handling the heterogeneous big data delivered by EO satellites and social media. The seemingly unrelated remote sensing science and daily social life happen to coincide by their nature as "big data." Many studies have shown that processing hundreds of thousands of online images and millions of online text messages is now possible [3], [15]. Consequently there is an impulse to develop a sophisticated system that effectively mines their information and coherently fuses them.

In this article, we discuss key aspects of geo-information harvesting from social media data, including social media data availability (Section II), social media data pre-processing and management (Section III), geo-information extraction from social media text messages (Section IV) and social media images (Section V), and the fusion of social media and remote sensing data (Section VI). Section VII showcases exemplary geographic applications. For the first time, we also extensively discuss ethical consideration of research with social media data (Section VIII) in the context of geo-information harvesting and geographic applications. Our aim is to inspire more researchers to explore social media data as an unconventional data source of geo-information, and provide a good basis for doing so.

## II. Social media data availability

Social media data is the information collected from social networks representing how users share, view, or engage with internet content and with each other. It is mainly composed of large quantities of photos, videos, and text messages, but also exists in many other forms, such as emoticons, product ratings, volunteered geographic information, and so on.

Despite the large quantity of social media data created every day, most of it is not available for geo-information extraction, mainly due to the license terms, the data crawlability, and the availability of geo-tags. Social media data is spread out on various online platforms that have different terms of use. For example, photos posted on Facebook usually do not have an open-access license, making massive crawling and processing of the photos on Facebook impossible. The same conditions apply to many other social media platforms, such as Instagram. In contrast, text messages posted on Twitter are by default open unless users restrict access; hence they can be crawled extensively. Twitter also provides an official API that permits massive download of tweets. Another issue is the geographic location contained in the data, as it is the key to linking the data to geographic applications. For this reason, this paper focuses on geographically harvestable social media data. We require that photos and text messages are either geo-tagged, or it is possible to infer their geographic locations with reasonable precision (in a building block level). Tabel I summarizes the available harvestable social media based on these s requirements. The list is of course not exhaustive. It is intended to give readers the most common sources of social media data for geo-information retrieval. Most freely available social media data is under the creative commons license, often the CC-BY license. The license of their images vary from platform to platform. We intend to give readers guidance on what type of social media sites and what data licenses are suitable for scientific research. For example, although CC-BY allows free use of the data, we note that a CC-BY license does not permit sublicensing. This means that posting or publishing such data on certain websites or journals requires special attention such as a sublicense or even the transfer of ownership, which is sometimes required by the journal publisher.

Our extensive research shows that among all these options, Twitter and Flickr are the most generous in terms of the total data volume and data harvestability, because of their license terms and the functionality of the APIs. *Volunteered geographic information* such as OSM is also accessible in large volume. However, they are not strictly social media data, hence not the focus of this paper. We will focus on tweets and Flickr images in the following.

**Twitter**: Twitter offers several API packages with different pricing levels. In this article, we focus solely on the freely available Twitter API, which allows the user to stream approximately 1% of the daily Twitter stream of an area of interest (AOI). Further, the API offers several techniques to query tweets. If a researcher is interested in a special hashtag[1] or keyword like #COVID19, it is possible to submit such a request to the API, and receive tweets with matching contents as a result. An AOI can be specified via a bounding box of a city, region, or country. It is also possible to receive the 1% stream without any keyword or hashtag filtering. A further method to receive tweets is to download a user's timeline or "hydrate" tweet IDs, i.e., retrieve tweet content based on IDs. This is particularly relevant in research as data sets are usually only shared in this format due to Twitter's license terms prohibiting sharing of complete tweet data.

**Flickr**: Flickr, a social media image platform, offers a powerful free API allowing arbitrary spatio-temporal queries. It can therefore be used to create comprehensive worldwide data sets, and is therefore often the first choice for image data. Flickr images have been used in various studies across all disciplines, e.g., [16]–[19]. In contrast, Facebook used to have an open API in its early days, but changed towards a more privacy-preserving one as it became more popular and allowed users more fine-grained options on the visibility of posts.

**Others**: Instagram closed their API in April 2018 and redesigned it to be used by businesses and external apps to curate the profile of a user. All access to query data specifically and randomly by time, location, or tags has been removed. Snapchat's API never offered any query features but targeted creation and publication of advertisements. Beyond these popular examples, there are several other platforms providing user-contributed images with geo-reference. Among them are *Google Places* and *Foursquare*, databases of points-of-interest with user images and reviews, *Geograph*, a platform systematically acquiring landscape photos across Great Britain and Ireland, and *Mapillary*, which aims to build a catalogue of street view imagery based on volunteers driving around with their own cameras. Until November 2016, *Panoramio* was a valuable source of geo-tagged images on a global scale and its data was used for several studies (e.g., [20]). This service has since been shut down, but its imagery is still available as a part of Google Maps.

Our recommendation is that readers use Flickr and Twitter for large-scale applications. Mapillary, Google Places, and Foursquare are good for study of specific locations. All of them provide official APIs.

## III. Social media data management

With respect to data management, a key aspect of social media data streams is that they represent a source of big data. Of the defining properties of big data – volume, variety and velocity [21] – velocity is the most obvious. However, the other two facets are also observed when working with location-based social media streams. The high velocity at which messages are generated leads to a huge number of messages, leading to significant volume. While these messages are often small, they may have bigger data assets associated with them, including images and videos for increased volume. Furthermore, there is a high degree of variety, from a technical point of view, in terms of the nature and quality of the location, time, and other metadata, from image and attachment file formats and semantics. The variety is also high from a user perspective. Including messages from professional marketing agencies, ethically sound information dissemination bots, bots actively engaging in specific topics to produce an impact on the perception (which can be considered unethical), users tweeting with their "professional personality," users using the network in a "private" setting, together with the network role and weight of a user (e.g., influencer vs. small network user), and even mixtures of all of these.

---

[1]Hashtags are keywords to tag a tweet with a certain topic, event, celebrity, etc. and consist of a word or word sequence without trailing spaces and a leading "#". For example #RemoteSensingIsGreat.

TABLE I
AVAILABLE SOCIAL MEDIA DATA

| Platform | Type | Description | License | Crawlability | Geotag |
|---|---|---|---|---|---|
| Google image search | Image | search engine | Partly CC or free to use | Third-party crawlers | partly |
| Flickr.com | Image | photographer website | Partly CC | Official API | partly |
| Unsplash.com | Image | photographer website | All CC or free to use | Official API | mostly |
| Pexels.com | Image | photographer website | All CC or free to use | Official API | mostly |
| Magdeleine.co | Image | photographer website | All CC0 or CC-BY | Official API | few |
| Twitter.com | Text and image | social sharing website | posts public by default | Official API 1% of all | partly |
| Instagram.com | Text and image | social sharing website | posts public by default | Third-party crawlers | partly |
| OpenStreetMap.org | vector and text | Online map service | Open Data Commons Open Database | Official API and DB dumps | all |
| Mapillary.com | Image | Online map service | CC BY-SA | Official API with limited quota | all |
| Google Places | Text and image | POI service | Proprietary | Official API | all |
| Foursquare | Text and image | POI service | Proprietary | Official API | all |
| Geograph.org.uk | Text and image | Online map service | CC BY-SA | Official API and DB dumps | all |

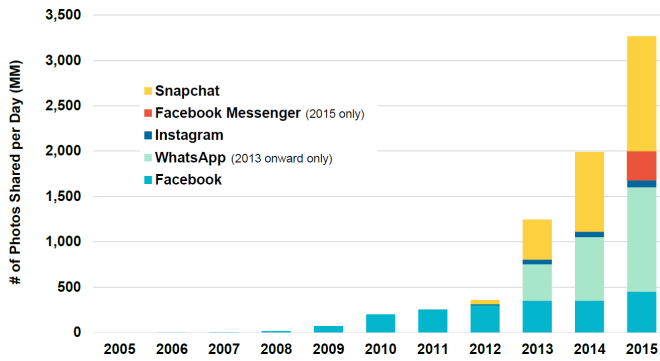

Fig. 1. Daily number of photos updated on selected platforms (© KPCB).

From a data management perspective, it is first most important to handle the big data characteristics. Distributed systems for taking care of data management are necessary. This implies that we can have only a single consistent key [22], and in fact infrastructures using key-value data representations dominate the field. With respect to key-value stores themselves, readers are referred to [23], [24].

A key in this context can be either a number or a short string. Depending on the use case, choosing these keys may require considerations of data locality versus random hash functions, which has implications for data access speed and node usage balancing.

For some geospatial applications on a global scale, query distribution will be quite uniform (e.g., generating a worldwide map) where data locality can be fruitful; for others it will be extremely local (e.g., disaster response) and random data distribution is a better choice. For example, if all data from the area of New York can be found in only few nodes, answering a single query about New York is efficient as only these few nodes need to be asked and coordinated for a definitive answer. However, if the majority of queries are relevant to the New York area only, then from a big distributed system only a few nodes can contribute and will become a bottleneck. Therefore, an optimal system needs to be designed with both data distribution and query distribution in mind.

The keys themselves can be built by a combination of random information, time, location, topic, hashtags, and other criteria. Spatial information can be integrated either as a set of keys (e.g., rectangles) or through the mechanism of space-filling curves. Such curves enable us to approximately map between 2D, 3D, or 4D space and a plain integer key which is used for ordering. In [25] an exemplary keying scheme based on message timestamps and a hash encoding of geolocations is presented. The locality of this pure spatiotemporal scheme is then reduced by introducing random characters.

In summary, managing social media data entails spatio-temporal data management as well as management of big data. Therefore, key-value structures are used to guide the low-level organization of the data and of the mapping of data to nodes. In current cloud computing infrastructures as well as in HPC-environments, a good dose of randomness needs to be inserted to avoid hotspots when managing a global dataset for local queries or local data for global queries.

## IV. GEO-INFORMATION EXTRACTION FROM SOCIAL MEDIA TEXT MESSAGES

As discussed in section II, Twitter is the most salient social media source with a focus on the text domain. In this section, we discuss various aspects of extracting geo-information from this source. In the past decade, Twitter has developed into a major service for sharing small texts which are called tweets (see Fig. 2). A tweet can include a news headline, an open position within a company, traffic information for a city, tomorrow's weather, a web URL to an interesting website, or more personal messages like feelings or opinions. All this information must be packed within 240 characters.

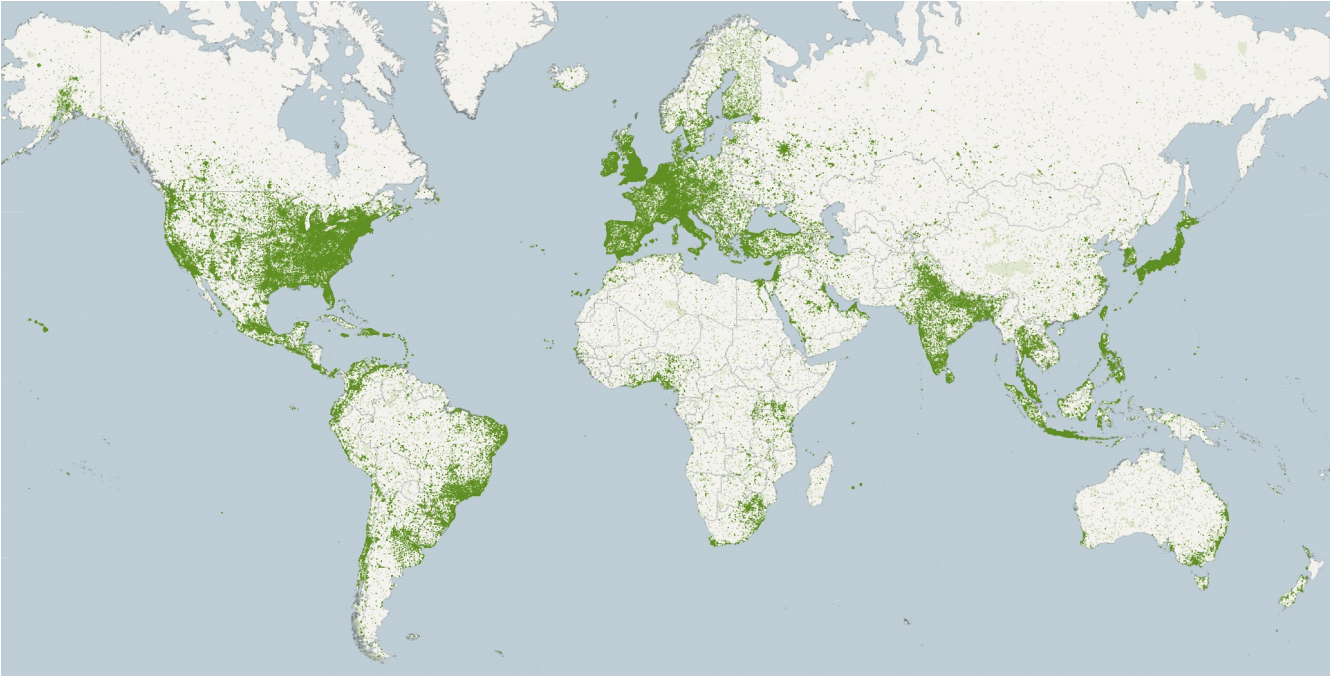

Fig. 2. Two-month world-wide Twitter sample with approximately 3M geo-referenced tweets.

A tweet can be analyzed to determine whether it includes (implicit) information about the surrounding environment and thus, can reveal further data about nearby urban characteristics, demographic information of a region, or events in the neighborhood. Due to the massive amount of data, automatic methods from the field of Natural Language Processing (NLP) are necessary here. This section gives an overview of such methods commonly used for geo-information extraction from social media material [26]. The algorithms and techniques shown here are widely language independent (or alternatives are depicted) and are applicable to all text-based social media. A key challenge when using Twitter data for geo-information extraction is the attachment of geolocations to tweets. The last part of this section discusses direct availability of such locations as well as solutions for obtaining tweet locations in other ways.

*A. Twitter Data Format and Pre-processing*

Tweets from the Twitter API are usually encoded as *json* [27] objects, which include several attributes. One provides information about the poster, including the user ID, user name, user language, a user description, sometimes the hometown, etc. Data about the tweet itself is also attached, such as the original and unprocessed tweet text. Further attributes are the tweet ID, the estimated language, a timestamp, a human-readable time of creation (UTC), and many others.

Before feeding the collected data into NLP or machine learning algorithms, it needs to be pre-processed because the raw social media text most likely includes informal spelling, typos, and creative use of punctuation like emoticons or emoji.[2] Common pre-processing steps include deleting URLs and user names, stripping punctuation marks and numbers, removing stop-words (*and, the, a, ...*), setting all characters to lowercase, and removing emoji. Of course, the choice of these methods depends on the use case. Subsequent pre-processing steps may involve lemmatization or stemming which convert all words into their roots, or normalization where irregular spellings like *yeeeeeeees* are corrected using simple rules or lexical knowledge. Widely used NLP libraries like spaCy [28], NLTK [29], or Gensim [30] offer implementations of these methods.

Many common techniques are particular to space-separated languages. For others like Japanese or Chinese, different approaches are necessary. For example, the Python library *Janome* [31] can be used to tokenize Japanese strings. It makes use of the MeCab dictionary [32], [33] including the Japanese new era (Reiwa) dictionary. For the Chinese language, *jieba* [34] has proved a useful tool for tokenizing Chinese text sequences.Some downstream NLP methods may require their own preprocessing steps, such as BERT (see section IV-B).

The methods discussed here mark a good starting point to pre-process Twitter, Weibo, or other (social media) text. However, there is no golden rule for text pre-processing and it always depends on the algorithms used for the desired task.

---

[2]Emoji are ideograms and pictograms depicting smiley faces with different sentiments, fruits, activities, items, or flags.

*B. Methods*

*1) Information Retrieval:* Aside from deep learning algorithms (which are discussed later), there are several "classic" information retrieval algorithms that are based in statistics. A widely used unsupervised algorithm to assign documents to certain topics in text documents is latent Dirichlet allocation (LDA, [35]). A topic can be seen as a word pattern that occurs in several documents (e.g., in a string or a tweet) and is represented as a bag-of-words. Documents with the same or similar word patterns are assumed to be related and therefore are clustered to a certain topic. Recently, [36] introduced an extension to LDA called Archetypal-LDA (A-LDA) which specializes in short texts like tweets using anchor words. Anchor words can be seen as a seed to "guide" the LDA topic inference. Hashtags were used as anchor words to work out topics around certain hashtags and therefore support the topic inference. This method could be useful in geo-spatial research, e.g., event detection, where certain hashtags are related to a specific event. Tweets without that hashtag could therefore be utilized as well.

LDA has been used in various studies in the geo sciences. For example, [37] used tweets and Flickr images for a multi-label land-use classification in New York and San Francisco. They applied LDA to extract relevant topics related to Foursquare venues in specified clusters. The relevant clusters were calculated by HDBSCAN [38] to identify local hot spots in the Flickr images. Before they applied LDA, the text was pre-processed by removing stop-words and applying lemmatization. Only those tweets within the clusters detected before were used. LDA revealed the relevant topics for each cluster, which led to good classification performance. LDA was also used in [39] to identify relevant topics of the Olympics Games 2012 in London within the context of city planning. First, the tweets were pre-processed with the aforementioned methods; thereafter, they used LDA to filter the tweets that were about the Olympic Games and transportation. Tweets that were related to the mentioned topics were used to perform a sentiment analysis and a spatial-temporal analysis. [40] investigated the tracking and monitoring of Twitter topics related to a disaster over time. Since LDA is not suitable for tracking topics over time, they used a dynamic topic model [41] (DTM), which is based on LDA.

Another classic, but still very popular [42] algorithm is term frequency-inverse document frequency (TF-IDF) [43]. Basically, TF-IDF measures the relevancy of a term (e.g., a word) in a document and weights the document's importance, for example for a search query. This weighing process involves the combination of two different steps. First, the standard way of determining the term-frequency (TF) of a certain term $t$ in a document $d$ is to calculate the quotient of the total count of $t$ appearing in $d$ by the complete count of all terms in $d$. Second, a stop-word like "the" would distort the document-weighting because it is likely to appear very frequently within a document. Therefore, the inverse document frequency (IDF) decreases the importance of the frequently recurring terms. This is achieved by taking the logarithm of the quotient of the total number of documents and the count of $d$ including that certain term $t$. The final TF-IDF score for a term is computed by the product of TF and IDF.

*2) Sentiment Analysis:* As pointed out in Section II, social media data can also contain emotions. One rule-based algorithm to determine sentiments in English (but not limited to) Twitter text data is Valence Aware Dictionary for sEntiment Reasoning (VADER) [44]. The authors compiled a list of sentiment-loaded terms. Those terms can be words like "happy" or emoticons like ":-)". Furthermore, the authors added support for emoji sentiment detection. The terms are validated and ranked by humans (Amazon Mechanical Turk). In the end, the list comprises 7520 of such terms. Every term in the list comes with a mean sentiment *intensity* ranging from $-4$ (very negative) to 4 (very positive) and ten independent human ratings. Next, a qualitative approach was used to detect the main textual drivers of the perceived sentiment intensity. With this, five heuristics incorporating grammatical and syntactical clues are derived to determine the sentiment intensity of a string or possible changes of the sentiment. For example, exclamation points add some sentiment intensity to a string, as does using all-caps to stress words that express the intended sentiment. Furthermore, booster words like "very" also contribute to the computation of the sentiment intensity. VADER achieved good performance and was able to outperform human reviewers in some cases.

*3) Embeddings:* Today, the field of Natural Language Processing (NLP) has seen a strong shift towards machine learning, and particularly deep learning, approaches, which in many cases now outperform previous methods based in linguistics [45]. The topic of social media analysis is no exception here, and most large-scale systems now employ deep neural networks [46].

A crucial issue that caused the relatively late introduction of deep learning to text data analysis (as opposed to other forms of data, like images) lies in how to represent words numerically. Traditionally, this has been done with one-hot encodings that do not capture semantic meaning. The key development here was the introduction of word embeddings. These embeddings are neural networks themselves, and are part of the complete classification network. Some very successful early approaches that are still in use today are *word2vec* [47], *GloVe* [48], and *fastText* [49]. One of the earliest, *word2vec* is a neural-based architecture that can embed large vocabularies from huge text corpora into an n-dimensional feature space very quickly. It can preserve semantic and syntactic features of words and embeds similar word close together in the feature space. This development was followed by *GloVe*, which is a statistics-based approach. To solve the issue of out-of-vocabulary (OOV) words, [49] introduced the *fastText* algorithm, which divides words into subwords. This process facilitates the approximation of OOV word vectors by composing a word vector based on the word's subwords, which is particularly useful for irregular text as it is often found on social media. In each case, it is very common to use pre-trained embeddings, such as those provided for *fastText* in 157 languages [50].

Urban areas are multilingual spaces [51]–[53] and the set of languages discovered in social media posts in cities is diverse [54] (of course, English is dominant on Twitter [55] and Chinese on Weibo). However, multilingual approaches are rare in earth sciences and so offer interesting research opportunities, such as ensemble models covering all languages to classify building functions in urban areas [56] or investigating the information density of Japanese or Chinese social media postings with respect to English [57] within the context of urban land use tasks.

As applied in [58], embeddings that represent not just words but whole sentences are also becoming used more widely, e.g., the Universal Sentence Encoder (USE) [59]. Here, the sentence embeddings are trained using either a Transformer-based approach [60] or a deep averaging network (DAN) [61]. In earth sciences, [62] showed the applicability of multilingual sentence embeddings. The researchers used a multilingual variant [63] of the universal sentence encoder [59] to conduct a sentiment analysis on geo-referenced European tweets related to the COVID-19 pandemic by fine-tuning the pre-trained sentence embedding with tweets. The application of multilingual sentence representations to classify sentiments by using a simple feed-forward neural network enhanced the results compared to a monolingual baseline. Of course, embeddings adapted specifically to tweets also exist, e.g., [64], [65].

In the past two years, BERT embeddings [66] and their various offshoots, which take complex contexts into account, have become very popular. BERT language models are based on the Transformer architecture [60] and are trained by using the so-called *masked language task* where a certain percentage of words in a sequence is *masked*. During training, BERT's goal is to predict the masked words, which increases context "awareness" of the model. BERT has been applied in studies focusing on urban areas such as sentiment analysis on energy-related complaints on Twitter [67] and the classification of flood-related tweets in Indonesia using the multilingual version of BERT [68].

*4) Neural Networks:* Embeddings usually serve as the input layer to a deep neural network, which can then be trained to solve various tasks like classification of tweets. Recurrent Neural Networks (RNNs) are commonly used for sequential data like texts, but due to the short-context nature of tweets, Convolutional Neural Networks (CNNs) are often more successful here and easier to train, with the architecture presented by Kim in [69] being used frequently. CNNs are also suitable for processing text that is represented at the character level [70]. Challenges like different languages or misspellings can be approached when working at a character level since no intuition of the network about words, semantics, or syntax of individual languages is needed [70]. For example, [71] shows the applicability of the character level approach by generating an English tweet and achieving good performance in various Twitter related tasks such as sentiment categorization. As pointed out earlier, since this approach is language independent, it could be an interesting technique for text classification tasks including social media texts in character based Asian languages such as Chinese or Japanese. It should be pointed out that most of the tasks described in the above sections about classical algorithms can now also be solved with neural networks, e.g., sentiment analysis [62].

Further usages of neural network-based approaches are presented in section VII.

*C. Geolocations of Tweets*

As discussed above, nearly all geo-related applications of Twitter data require information about the geolocation where each tweet was posted or refers to. Around 1% of all tweets are already geo-tagged explicitly [72]: that is around 500M [73] tweets are published per-day, and 5M of them are geo-tagged. Each geo-referenced social media post could easily be resolved/decoded to a specific location on Earth. By aligning the geo-referenced social media content with an openly available Geographic Information System (GIS), such as OSM, we have a valuable, easily accessible, and cheap source of information. In this context, social media data contributes to Volunteered Geographical Information (VGI), and, consequently, empowers "citizens as sensors" [74], [75].

When working with geo-tagged social media data, we can differentiate between two types of geo-locations. The first is the geo-location of the person/app who posted or created the content, for example, the GPS coordinates of the phone from which a tweet is posted. The second is the geo-location that is cited within the social media post, e.g., a Point Of Interest (POI). While the second is increasingly supported by social media channels, the first is less and less supported, ostensibly due to privacy issues. Twitter stopped supporting the first type of locations in mid-2019 [76], [77], but still supports the second one via different mechanisms, like explicitly tagging a tweet using one of the nearby "Twitter Places", or implicitly mentioning a POI within the content of the tweet.

Obtaining exact/named geo-locations from social media is a challenging issue, especially the precise location of the person when they create a post on social media. As mentioned, this kind of location is unsupported in social media apps/sites and users normally do not give social media apps the right to access their location to protect their privacy. Moreover, a tagged POI is potentially useless because of its coarse granularity, as in most cases users tag a country or a city (see Table II). In addition, the implicit mention of a POI in a social media post poses its own challenges. In textual posts, we need a mechanism to identify named entities and then resolve them to a precise location. In visual posts (image or video), the challenge of identifying the POI that appears in the scene is even greater. Above all, the volume of available data, and the fact that social media data is unstructured, heterogeneous, and noisy, makes social media a challenging source of information [75]. Based on our experience with geo-referenced Twitter data, the main data challenges can be summarized as follows.

*a) Precise Locations are Unsupported:* In tweets, precise locations refer to the geo-coordinates of the person/app who

TABLE II
GRANULARITY AND SHARE OF TAGGED PLACES OF GEO-TAGGED TWEETS COMING FROM NATIVE TWITTER APPS. NATIVE APPS ARE: "TWITTER FOR IPHONE", "TWITTER FOR IPAD", "TWITTER FOR ANDROID", AND "TWITTER WEB CLIENT".

|  | Country | City | Admin | Neighborhood | POI |
|---|---|---|---|---|---|
| April 2019 | 3.9% | 84.3% | 10.1% | 0.1% | 1.6% |
| May 2019 | 3.8% | 84.4% | 10.1% | 0.1% | 1.6% |
| July 2019 | 3.8% | 84.2% | 10.6% | 0.1% | 1.3% |
| August 2019 | 3.5% | 84.6% | 10.4% | 0.1% | 1.4% |
| September 2019 | 3.3% | 84.8% | 10.4% | 0.1% | 1.4% |

TABLE III
SHARE OF TWEETS COMING FROM NATIVE AND THIRD-PARTY APPS AMONG ALL GEO-TAGGED TWEETS. NATIVE APPS ARE: "TWITTER FOR IPHONE", "TWITTER FOR IPAD", "TWITTER FOR ANDROID", AND "TWITTER WEB CLIENT".

|  | Native apps (%) | Third-party apps (%) |
|---|---|---|
| April 2019 | 84.1% | 15.9% |
| May 2019 | 84.7% | 15.3% |
| July 2019 | 85.7% | 14.3% |
| August 2019 | 86.2% | 13.8% |
| September 2019 | 87.2% | 12.8% |

created the tweets, at the creation time. According to a Twitter announcement,[3] this type of location was well-supported before mid-2019, but has not been supported since then. This seems to be a general tendency in the other social media channels as well, to better protect the privacy of users. This restriction already prevents many geo-information applications from using social media data. For example, the building function classification downstream task, which enables the function of a building (e.g., commercial, residential, etc.) to be determined based on the topics referenced near it, requires the precise geo-location of each tweet. Consequently, the only remaining type of supported geo-locations in Twitter data is the geo-location of hot spots, such as POI, neighborhood, city, country, etc., which are explicitly mentioned by users within the tweets or implicitly assigned by the Twitter app. In fact, even before mid-2019, the number of tweets that were precisely geo-tagged and that came from the native Twitter app accounted for only 8% to 25% of all geo-tagged tweets [76].

*b) Arbitrary Coordinates and Insufficient Metadata:*
Each tweet consists of textual content and metadata, where metadata is used to encode information about the tweet, such as user ID, geo-coordinates, place type and name. The Twitter website and Twitter native apps provide fairly rich metadata for each tweet. However, a considerable share of all geo-referenced tweets, around 15% (see Table III), come from third-party apps like Instagram and Foursquare; for these tweets, metadata is either missing or inaccurate. To sum up, although there are many geo-referenced tweets, in many cases the exact type of those coordinates cannot be detected; that is, we not know if they refer to precise locations or to certain POIs.

[3] https://twitter.com/TwitterSupport/status/1142130343715078144?s=20

*c) Granularity of Geo-locations:* It is not only the availability of geo-referenced social media data that makes the difference, but the granularity of their related coordinates. For example, to classify buildings or to identify hot spots in a city, we need geo-coordinates on a scale fine enough to identify an individual building. After exploring Twitter data, we found that most of the tweets are assigned city- or country-level coordinates. It seems that if the user does not specify a location, the platform fills the metadata field with the user's city- or country-level coordinates (see Table II). In addition, a considerable number of tweets are coming with polygon coordinates rather than a point coordinate, and the polygon is sometimes too big to extract useful information.

Before proceeding with research that depends on such geo-tagged data, we need to first consider two questions. First, are the geo-locations that we need available? And if so, are they of acceptable quality? Second, what is the available granularity of geo-locations (e.g., building, neighborhood/polygon, city, country)? Concerning the first question, it seems that social media sources are increasingly providing POI locations instead of the exact location of the person who creates the post. For the second question, researchers need to focus on an acceptable level of granularity, i.e., avoiding fine-granularity geo-tagged location like buildings coordinates. Therefore, research design needs to consider the different possibilities of increasing the quality of the available data, looking for new sources of data, or re-adapting the downstream applications to be compatible with what the data offers.

As mentioned earlier, more than 99% of tweets are not geo-tagged, which means that most tweets, in their original state, are not usable for geo-applications. On the other hand, it means that there is a great opportunity to increase the amount of available geo-tagged data if these tweets [78] could be geo-tagged, as in the example in Fig. 3. Fig. 4 shows that using a pre-trained basic NER algorithm [79], we are able to identify "Location" entities in 6% of all non geo-tagged tweets. Furthermore, we were able to identify "Organization" and "Person" entities in 13.5% and 13.3% of tweets, respectively, where many of "Organization" and "Person" entities could be geo-encoded to a certain location (see Fig. 3).

V. GEO-INFORMATION EXTRACTION FROM SOCIAL MEDIA IMAGES

To gain spatial knowledge from social media images, they must have a geo-tag that allows a precise localization of where the image was taken. Most social media platforms enable their users to upload images and set a location to let others know where their images were taken. A set of such localized images can provide deep insights into the surrounding area where they were taken, e.g., activities, landmarks, and land cover. Extracting knowledge from this vast variety of images requires a structured approach: Fig. 5 sketches a general pipeline, from data acquisition to generating machine learning-driven models, for extracting geo-spatial knowledge.

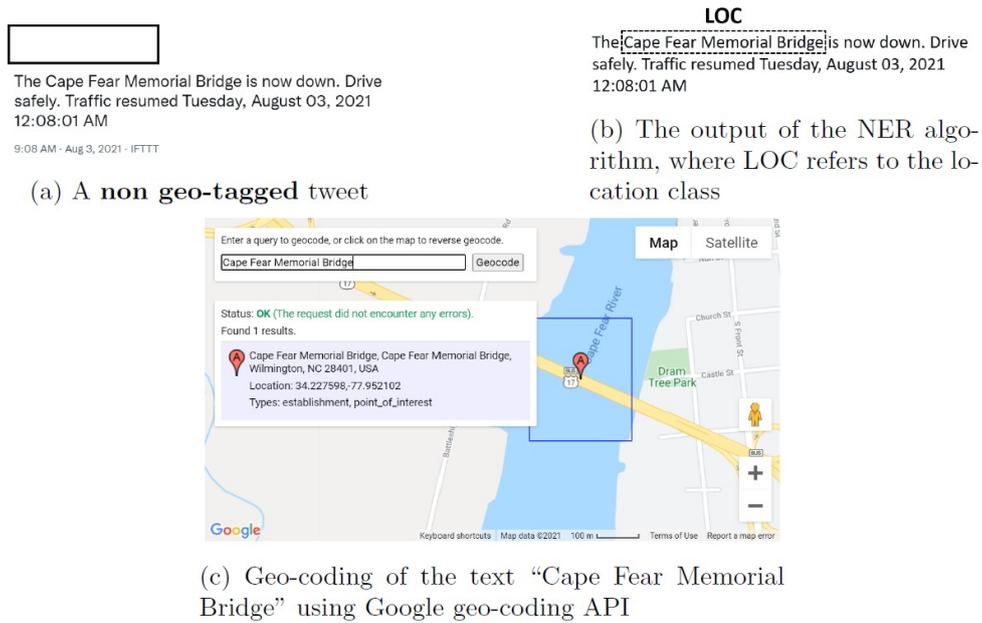

Fig. 3. NER for geo-coding of non geo-tagged tweets.

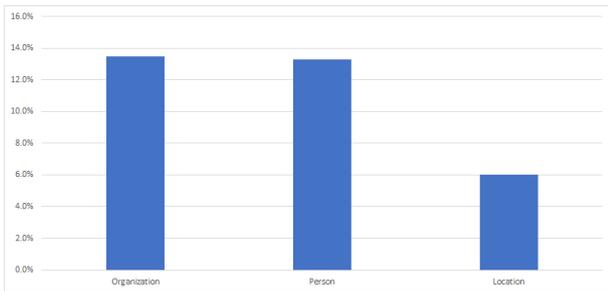

Fig. 4. The share of identified NER-tags when applied to the non geo-tagged tweets that are collected between August 3 and 9, 2021.

### A. Social Media Image Pre-processing

While most social media platforms let users tag their photos with a location, external access to this information is limited. Furthermore, the accuracy of these geo-tags must be treated with care, since manual tagging is error-prone and GPS sensors have limited accuracy when signals are distorted [80]. If an image is taken with a smartphone camera it will automatically be tagged based on the signal from a built-in GPS sensor. Otherwise, geo-tags can be entered manually while creating the social media post. In this case photographers tend to tag batches of images and put their geo-reference to the neighborhood where they were taken instead of a precise location, which is presumably unknown.

Manual tagging also involves pictures that share no clear relation to the location. Especially art-related images with geo-reference do not show anything about the surrounding area where they were taken, but draw the attention to a certain motif. Due to their vast variety, finding a subset of images that is helpful for a given task is crucial.

To cope with these intrinsic issues, feature extraction and filtering are at the core of every methodology to extract geo-information from social media images. An initial filtering step can be based on visual content screening [81] or keywords from the metadata of each image [82]. Moreover, pre-trained models for image classification and object detection help in understanding the image contents and eliminate the need to train on data with uncontrollable quality. Models trained on the ImageNet classification task have been successfully applied to obtain general and abstract image features from hidden layers of neural networks [83]. Object detection models provide an intuitive insight into items present at certain locations, which can be aggregated on a spatial level afterwards to classify urban land use [84].

Putting the extracted information into the spatial domain is mostly done by aggregating it into raster-based formats. This eliminates smaller errors from GPS sensors and set the information into a larger spatial context. On the other hand, this approach might lead to vanishing information if signals from different images are highly diverse.

Relating objects in images to spatial objects requires more than a geo-tag. If a geo-tag and a compass direction are present, a line-of-sight can be calculated and related to the spatial objects that it intersects [85]. More sophisticated approaches can make use of the camera's angle-of-view to define the area depicted in an image. However, as the standard for metadata does not take into account information about roll and pitch, such algorithms have to assume that the camera's position was nearly orthogonal to the ground.

### B. Methods

Geo-referenced social media images have been used for land cover and land use classification on various spatial levels, from the parcel down to the building instance level [83], [86]–[93]. Additionally, they can be used to automatically identify landmarks and points-of-interest due to spatial clusters, resulting from multiple users taking photos in one area with the same motif [94]. Zhu et al. [93] use Flickr images to predict one of 45 pre-defined land-use classes on a parcel level. They train their model on global Google and Flickr weakly-labeled images, testing with Flickr images in San Francisco. They jointly fine-tune two CNN networks. The first network, called the object stream, is a ResNet101 architecture pre-trained on ImageNet, in which the last layer is replaced by a 45-class classification head. The second network, called the scene stream, is also a ResNet101 architecture pre-trained on the Places365 dataset, in whcih the last layer is also replaced by a 45-class classification head. The final classification decision is a late average fusion of the outputs of the two sub-networks. Hoffmann et al. [83] address the same problem as Zhu et al. [93] but at the more fine-grained level of the individual building. Hoffmann et al. use a classification scheme with 5 classes, and train and test their model using Flickr images in Los Angeles. They simply assign each geo-tagged Flickr image to the closest OSM building in Los Angeles, assigning the label of the building to the image. Then to classify an image, they first transform the image to a vector of features using VGG16, a state-of-the-art vision neural network pre-trained on ImageNet. Next, they pass feature vectors to a 5-classes logistic regression classifier. Finally, the building-level predicted class is the majority vote of the predicted classes of the images that are assigned to that building.

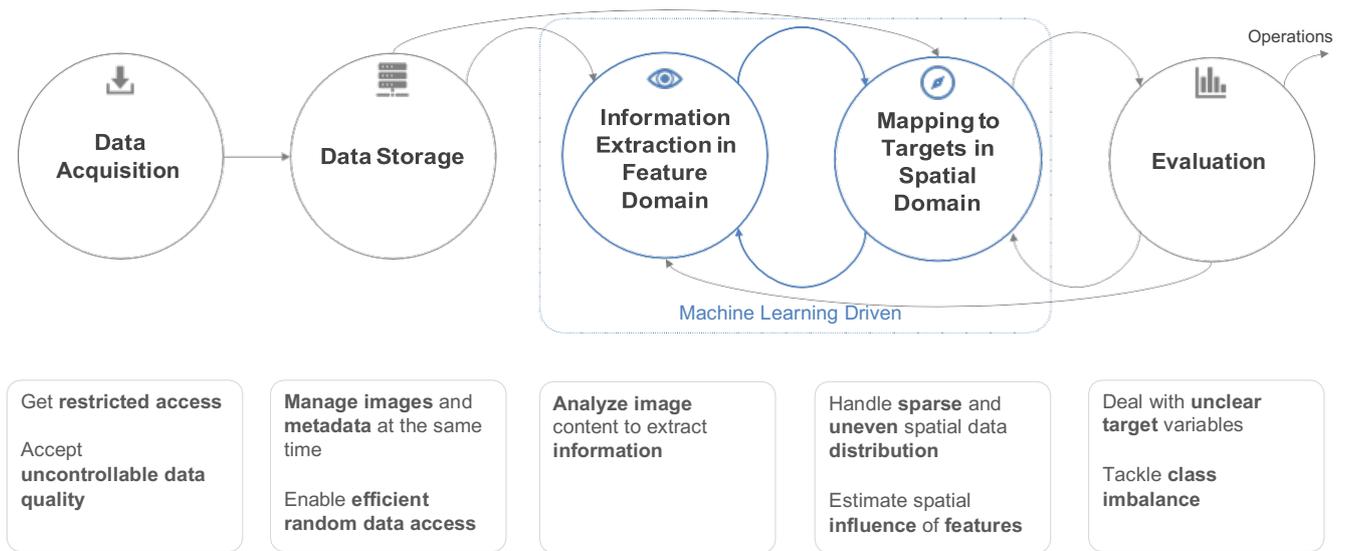

Fig. 5. Data processing chain for extracting geo-spatial knowledge from geo-tagged social media images.

Social media images can provide critical hints during natural disasters. There are two different datasets supporting this application: one focusing on image retrieval, i.e., finding all images related to a disaster in the stream of all incoming data [95], and the other aiming at image classification so that the correct situation, like flooding or fire, can be reported to emergency teams [96]. In the case of flooding, studies have shown that social media images can be used to estimate the water level [97], [98]. For example, Chaudhary et al. use multi-task learning on a dataset that is manually compiled from Flickr, Google, and National Geographic. As accurate water levels in meters are cumbersome to collect, their model learns to jointly predict water levels as a regression tasks and relative levels as a ranking task. The assessment of water levels is based on five common objects in urban environments: persons, cars, buses, bicycles, and buildings/houses, which allows the ground truth to be estimated by relative comparison. However, this method does not use any geotags for spatial interpretation.

Another application is environmental monitoring, either by mapping animal and plant species [99] or by assessing landscape preferences [100] and aesthetics [81], [101]. Urban areas can especially benefit from this data by collecting citizen sentiments and opinions from social media image content and textual descriptions [102]. For example, in a study of Copenhagen, Instagram images were collected by hashtag search and manually assigned to six categories of urban nature [103]. The researchers then performed a spatial analysis, where images of certain categories appear, and identified hot spots, where inhabitants of a city like to be, as well as what attracts them.

Relating urban green space information with health data does not show any connections [104], but social media images do contain information about citizens' habits. For example, they are suited to analyze conditions like alcoholism on a country level [105]. Beyond these insights, other studies have investigated more latent socio-economic variables like GDP, ethnicity, population density, and medical indicators [106]. Instead of trying to predict the exact geo-location of a given image, which is a hard problem to solve, especially in sparsely photographed places, Lee et al. [106] propose a CNN-based approach to predict 15 socio-economic attributes, such as GDP, infant mortality rate, and population density using Flickr images. They create a labeled dataset of Flickr images, in which each image is weakly labeled using at least one of the 15 attributes. Then, they fine-tune one of the pre-trained state-of-the-art ImageNet models by replacing the last layer with a binary classification head for each attribute. Finally, they have 15 binary classifiers, one classifier per attribute. For example, the GDP classifier predicts if the input image belongs/shows a place with high/low GDP, and so on.

Furthermore, social media images have received attention from the remote sensing community since these images provide a complementary view on the nadir perspective of satellites. Studies have shown that remote sensing images can be labeled using social media images [107], [108] or outcomes of remote sensing models can be verified with social media images [82]. The latter used a keyword search in Flickr to obtain a set of images that was used to verify the labels of GlobCover [109] in western California. Their verification is based on the classification results of a VGG16 network fine-tuned on a global set of Flickr images for the given task. They reported an overall accuracy of 83.80% of GlobCover from their approach, which is slightly higher than the human verification results of 80.45%.

Choosing a social platform depends on the use case. Most studies for disaster response rely on Twitter stream data, as this is a time-critical application requiring fast information flows. Flickr is best suited for analysis of spatial distribution, e.g., species monitoring or other tasks that are time invariant. While Instagram and Facebook used to be potential data sources before 2018, they are no longer applicable, as they closed their APIs for content crawling.

Recent advances in aerial-to-ground mapping are potential solutions to this issue [110], [111]. These methods aim at learning a joint representation of aerial and ground views that can be helpful for orientation and pose estimation without having a LiDAR-derived point cloud. Beyond this fusion of social media and remote sensing data, there are several other aspects of aerial-to-ground mapping that are covered in the next section.

*C. Geolocations of Social Media Images*

Ideally a given social-media image has accurate location and time metadata. When this is not available, it may be possible to estimate this information. One approach to estimating the location of a social media image is cross-view localization. In this setting, a ground-level image is localized by matching image features to features extracted from an overhead imagery source. Early work used traditional image features [112]. Subsequent work explored the use of features from deep CNNs, either frozen, pretrained networks [113] or networks optimized for the localization task [114], [115]. Since these earlier works were published, much work has focused on localization of panoramic images, leading to models that focus on the richer geometric models [116], [117].

A key component of localization is the ability to extract location-discriminative features from overhead imagery. Another line of work has focused on instead extracting semantic features that are derived from social media imagery, and other sources. These features may not be as discriminative of location, but are potentially useful for other applications. Some examples of mapping image features include mapping scene categories [115], the distribution of objects [118], and combinations of these features [119]. One potential application of this work is in using social media as a source of supervision for training overhead image understanding models, such as the work by Greenwell et al. [120] that highlights the relationship between object distributions and landuse categories. Beyond images, similar approaches have been used to map features extracted from geo-tagged audio [121].

Most work on metadata imputation/verification has focused on the localization task. This is in part due to the difficulty of obtaining sufficient training data. However, several works have attempted to estimate timestamps. Salem et al. [122] proposed a novel architecture for predicting object distributions and scene categories based on the location, time, and satellite image of the location. While their central focus was on mapping tasks, they demonstrated the ability to estimate a probability distribution over the capture time. More recent work, by Padilha et al. [123] has significantly improved the performance on both the problems of verifying whether a purported timestamp is correct and estimating the distribution over possibly valid timestamps.

VI. FUSION OF SOCIAL MEDIA AND REMOTE SENSING DATA

Traditional remote sensing imagery is collected by satellite and airborne platforms. This view provides a large potential for applications, however, with limitations due to capturing only physical conditions on the Earth's surface or in the atmosphere, infrequent collection, and delays in availability. In contrast, social media data is not as systematic and standardized as remote sensing data, but it is collected frequently in many areas, is often available within seconds of when it was collected, and provides views not possible from overhead platforms, both indoors and of exterior walls.

The key advantage of remote sensing imagery is that it typically densely samples a given area so that the geographic location imaged by each pixel is known, often with only a few meters of uncertainty. The collection process for social media is more variable and less reliable, since there is less control over the collection process. There is also a bias in the collected data, both due to the interests of persons and their interpretation of what they are seeing. Given their complementary nature, fusing the two data sources has the potential to help solve problems that are not addressable using remote sensing data alone.

*A. Fusion of Remote Sensing and Social Media Imagery*

When working with social media imagery and text, the their location is less precise and the orientation is very rarely known. Typically, the orientation is ignored and the features are extracted from the ground-level imagery by either averaging over wider areas [93] or using nearest-neighbor approaches [124]. This limits the value of the ground-level imagery to capturing coarse properties of an area.

Early work on fusing ground-level and overhead views focused on street-level panoramas, such as those in Google Street View. Unlike social media, street-level panoramas are collected using carefully engineered platforms to ensure consistent imagery and metadata quality. While still present, the location and orientation uncertainty is significantly lower than it is for social media data. The key unknown is the scene geometry, which means that directly relating ground-level image features to overhead imagery is a non-trivial problem. Study also shows that it is possible to use structure-from-motion techniques to rapidly construct large-scale static 3D models from social media imagery [125]. However, such models are typically restricted to areas with large numbers of social media images. More recently, Li et al. [126] demonstrated the ability to construct time-varying 3D models from social media imagery. When geographically registered, such models, both static and dynamic, have the potential to be integrated with remote sensing imagery using similar models to those used for street-level panoramas.

The known location and orientation has enabled approaches that rely on the known geometric relationships between the viewpoints, such as Workman et al. [127], which converts individual panoramas into oriented perspective cutouts and fuses them with satellite data using kernel-based feature fusion. Wojna et al. [128] propose an approach that projects features from the ground-level imagery into the overhead perspective by using building outlines, ignoring cues from the ground-level imagery, such as occlusions. More recent work by Workman et al. [129] proposes a geospatial attention model that enables flexible fusion of features from ground-level and overhead imagery. They demonstrate significant improvements over previous works on a variety of urban-area understanding tasks. Future work on incorporating such imagery will likely incorporate recent improvements in monocular depth estimation [130] that will allow more precise localization of ground-level image features in the overhead views.

More studies address the decision-level or feature-level fusion, instead of the more challenging pixel-level fusion mentioned above. [131] compared three different fusion models of aerial-view and street-view imagery for building function classification. Rosser et al. [132] propose a Bayesian approach combining information from Flickr imagery and LANDSAT-8 imagery to estimate flood inundation. Flood extents detected in the Flickr imagery are converted into the overhead imagery using viewshed analysis from a LiDAR-derived digital terrain model. Their results show that using social-media imagery alone leads to poor performance, but combining the two works better than using only the satellite imagery. This study illustrates how unknown viewing direction is one of the key challenges in working with Flickr imagery. Only four of 205 images include metadata (i.e., the GPSImageDirection EXIF tag). One possible solution would be to estimate the viewing direction and other calibration parameters, but this is a challenging problem, especially in regions with limited reference imagery. Instead, they assumed that the camera had a 360° field of view and that if flooding is present in the image, it is uniformly distributed in the viewshed. These simplifying assumptions assume that every Flickr image is a full panorama and that every imaged pixel is covered with water. While these are clearly invalid simplifying assumptions, it is impressive that even with these in place the social media imagery is able to improve performance over the use of satellite imagery alone. Assumpção et al. [133] provide a broad overview of the integration of social media, including imagery, video, and text, with hydrological models and remote sensing data for the general problem of flood modeling.

*B. Fusion of Remote Sensing Data and Social Media Text*

Social media text messages seldom contains precise location or orientation information like those in the metadata of some social media images. Therefore, the fusion of text messages and remote sensing data can only be done in a feature- or decision-level.

Hultquist et al. [134] present a case study on estimating the extent of a power outage following a major hurricane. They combine night light remote sensing imagery with the distribution of tweets that were tagged as "power related" by a machine-learning model to estimate areas where the power is out, providing street-level resolution. They find that the key benefit of the social media text is in increasing the spatial resolution of the model. Leichter et al. [135] use gradient boosting to combine features derived from remote sensing imagery and tweets to estimate local climate zones. In this approach, no advanced text understanding is performed. Instead, tweet features are simple values, such as total number of tweets in an area or the average tweet length. Including the tweet features improved the average F1 score from 49.9 to 53.1. In [136], the combination of tweets and remote sensing images improves the overall prediction results of both individual modalities [136] in building function classification, demonstrating the complementarity of the two data types. Additionally, the authors show with a spatial cross-validation that the models can generalize beyond a certain region. TEST Figure 6 depicts examples of classification results after the fusion process documented in Figure 7.

## VII. GEOGRAPHIC APPLICATIONS

The methodologies introduced in the previous sections have enabled various applications in geo-information retrieval. Where many people live, there is much communication. The manifold land uses, the contextual embedding of space into meanings and perceptions, and dynamic changes in urban space due to daily life routines can only be indirectly assessed by social media data. Studies show that social media data such as from Twitter, Flickr, or other geolocated posts has the potential to add the non-physical in objective or subjective form, e.g., [137], [138], biased by those who use these platforms [139], [140]. The most commonly addressed application is land use land cover classification, as images and texts can give direct information of the building appearance or the functionality. Urban geography and social science are also widely studied fields. The applications are diverse, including public health, predicting socio-economic variables, mobility, POI, and sentiment analysis, to name a few.

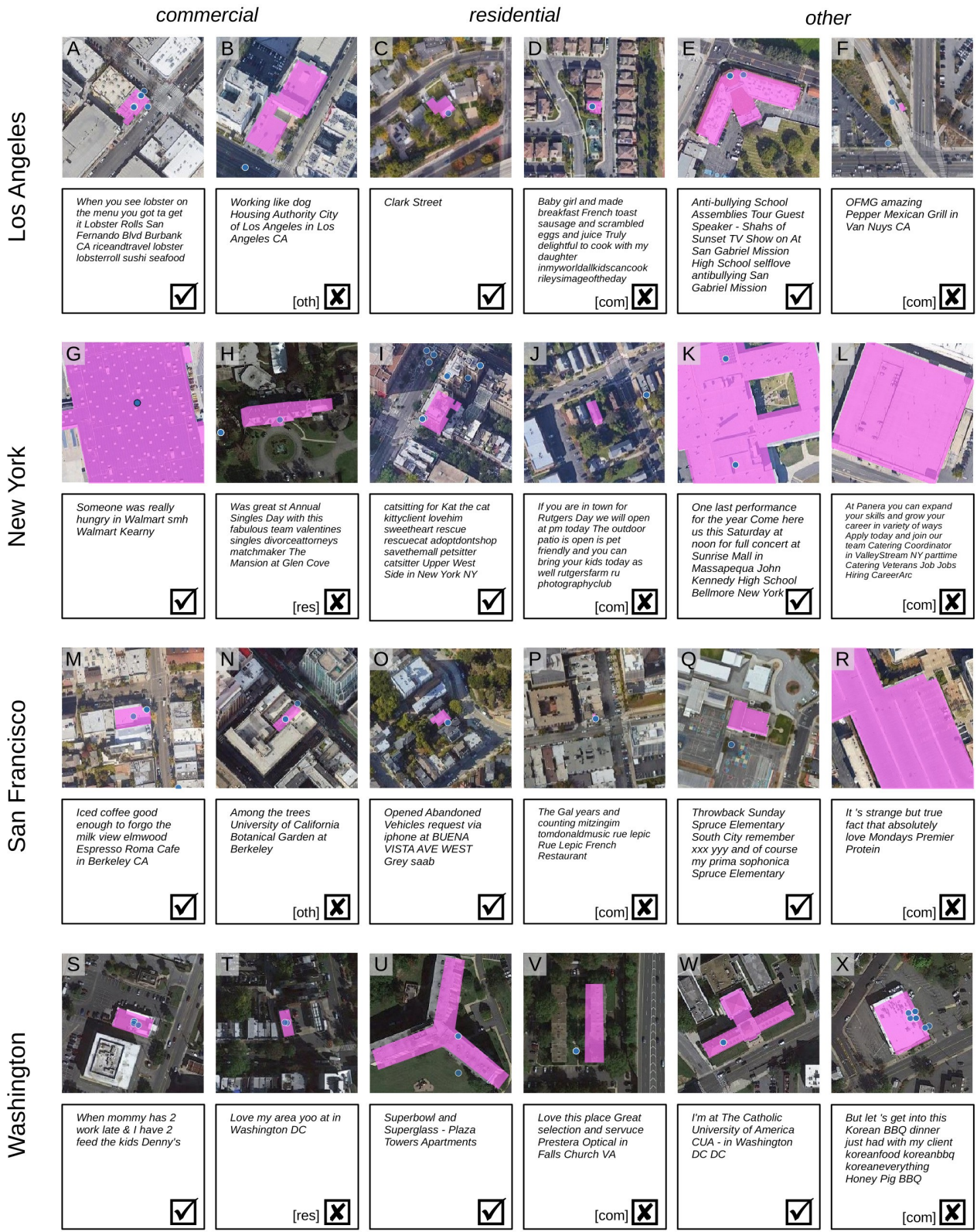

Fig. 6. Very high resolution aerial imagery with OpenStreetMap building footprints and corresponding Twitter data. The abbreviations in the brackets denote the wrongly estimated class (*com* = commercial, *res* = residential, *oth* = other). Real names were masked with *xxx* or *yyy* to preserve the privacy of the Twitter users. Image taken from [136]. Background images ©TerraMetrics 2021, Google.

TABLE IV
SUMMARY OF GEO-INFORMATION RETRIEVAL FROM SOCIAL MEDIA DATA.

| Geo-information | Text | Images | Datasets, Code |
|---|---|---|---|
| Land use land cover | [37], [58], [136], [141]–[145] | [37], [83], [85]–[93] | https://syncandshare.lrz.de/getlink/ fiFS5He9bZsR4Urh8hZGDGg/BIC_GSV.tar.gz [146] |
| Urban geography, social science | [147], [148] | [104]–[106], [148] | |
| Human perception, sentiment | [39], [138], [149]–[151] | [81], [99]–[103], [152]–[155] | https://github.com/abhimanyudubey/dlcity [152], http://scenicornot.datasciencelab.co.uk [155] |
| Crisis response | [40], [75], [156]–[163] | [95]–[98], [157] | CrisisLex T26 [159] [164], CrisisNLP [165], CrisisMMD [166], TREC-IS 2019A [167], CrisisTracker [168], https://www.ushahidi.com/, TREC Incident Streams http://dcs.gla.ac.uk/richardm/TREC_IS/ |
| Other resources | [28]–[34] | [115] | https://mocobeta.github.io/janome/en7 [31], https://spacy.io/ [28], https://www.nltk.org/ [29], https://radimrehurek.com/gensim/ [30], https://taku910.github.io/mecab/ [32], http://jaist.dl.sourceforge.net/project/mecab/mecab-ipadic/2.7.0-20070801/mecab-ipadic-2.7.0-20070801.tar.gz [33], https://github.com/fxsjy/jieba [34], https://mvrl.cse.wustl.edu/datasets/cvusa/ [115] |

In order to give a concise overview, we sort these applications into the following four topics: land use and land cover classification, urban geography and social science, environment perception and sentiment analysis, and crisis response. Although land use land cover classification may overlap with urban geography, we isolate it as a single topic, as it is the most commonly addressed problem. Table IV summarizes the applications, the associated literature, and resources. In the following, we provide an overview of the development in each application field. In each field, one detailed study will be given as an example.

*A. Land Use and Land Cover*

Land use and land cover classification are related, but cover different aspects. While land cover is measurable, land use requires the interpretation of data [169]. However, if social media data is used to predict this geo-information, both cases require the interpretation of patterns using machine learning algorithms. There are different aspects of studies in this area, including scale, data source, spatial granularity, classification granularity, and method. Concerning the scale, most studies are performed at the city level, e.g., [83], [88]–[90], [92], [93], with [85], [91], [136], [145] as notable examples of multicultural scales. Flickr is the most frequently used data source for images [83], [85]–[90], [93], whereas Twitter is mainly utilized for text applications [136], [141]–[145]. The spatial scale has a high variety: from grids of various sizes as in [89], [144], [145], to parcels [86], [87], [93], buildings [58], [85], [131], [136], and images [91]. Since there is no standard for land use classification schemes, a broad range of them has been proposed, from generic ones with three classes [85], [136] to very fine-grained schemes of 45 classes [93]. Early approaches used handcrafted features [86], [89], [92], while more recent methods have increasingly applied deep learning to solve the task [58], [83], [85], [87], [90], [91], [93], [136], [143]. Among them, some works address the fusion of ground-level and aerial images for land use land cover classification, for example [131]. Aerial image scene classification is a highly related but standalone topic. Readers could refer to articles in remote sensing [170]–[172] for more insights.

Social media images can latently encode information about both land use and land cover. A social media image may, for example, show a building with parts of the surrounding garden, thus containing land cover information like grass, and hints about land use in the building façade. The similarity of both classification tasks enables joint approaches with deep learning architectures [83], [90]. This co-occurrence of land cover and land use is found primarily in images, while social media text predominantly contains information about land use only because people use Twitter, for example, to tweet about their activities. Embedding models like *fastText* (see Section IV-B) can help detect hidden information by extracting abstract feature vectors. Used as a feature encoder to a bi-LSTM, *fastText* allows the accurate predictions of building functions [58]. Moreover, the combination of tweets and remote sensing images improves the overall prediction results of both individual modalities [136]. Figure 7 illustrates the pipline. To implement this, predictions are generated for individual tweets and then aggregated on a building level. Afterwards, this aggregated prediction is fused with predictions obtained from very high resolution images via CNN-based models.

With the evolution of multimodal networks that combine image and text seamlessly, we expect to see further approaches towards this direction in the future. However, since no benchmark datasets are available for this task, a comparison of different methods is only possible at a qualitative level.

*B. Urban Geography*

In urban geography, thematic applications from the fusion of remote sensing and social media data are manifold: examples include the mapping of population [173], the analysis of urban green space with public health [104], [105], the prediction of socio-economic variables such as GDP, ethnicity, population density, and medical indicators [106], [173], the assessment of cultural ecosystem services [153] or sustainable development [147], and the determination of house prices [174], among many others. The most obvious and widespread application, however, is the classification of building usages. A building's physical shell is not always clearly described in remote sensing data, but place, time, frequency, and content statements in social media data make it possible to assign uses such as office, residential, industrial or similar with high accuracy [141], [143], [175]. These variable uses also define the high spatial dynamics of people's daily routines. Social media data allows researchers to analyze activity/mobility patterns [142], [148]. People using smartphones or interacting with social networks are becoming "citizen sensors" [137], making it possible to relate static urban space to movement [176].

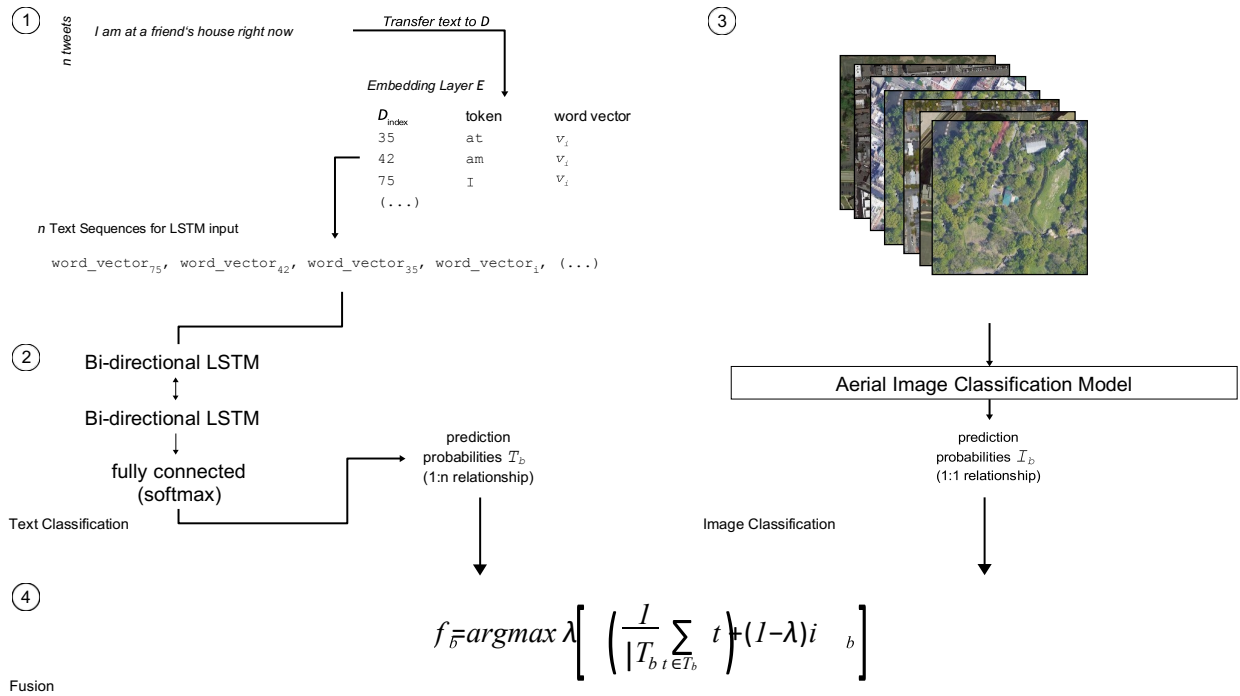

Fig. 7. Fusion framework from [136]. *1)* Mapping the text to machine-readable representation. *2)* Text classification with stacked bi-directional LSTMs. *3)* Remote sensing image classification with a CNN *4)* Decision-level fusion by weight-averaging the prediction probabilities of text and image classification. Aerial images ©TerraMetrics 2021, Google.

One successful application shown in [145] is the correlation between different population groups and their participation in social media communication in urban contexts. The frequency and times of tweets in certain urban areas are compared with those in other areas in order to determine the participation of these groups in modern communication. Fig. 8 shows an example in this respect: in [145], based on remote sensing data, the built-up urban landscape is divided into formal settlement structures and informal settlement, commonly known as morphological slum areas. The analysis of geolocalized Twitter frequencies over time then allows us to derive spatial deviations from the median of the respective city, into so-called digital centers (hot spots) or digital deserts (cold spots). By superimposing these datasets it is possible to determine that in morphological slums the participation in social network communication is actually lower. These approaches are comparatively objective, despite the bias of the basic population mentioned above. This study also revealed that among Twitter users, both sides of the economic divide exhibit very similar temporal behavior patterns.

In general, however, it must be stated that methods and applications in areas of tension between urban geography, remote sensing, linguistics, and other interdisciplinary fields are still in their infancy.

### C. Human Perception and Sentiment

The two previous subsections primarily discussed land characteristics observed through categories meant to be "objective." The debate of the universality of these categories notwithstanding, their study can be pursued using the ever growing amount of labeled data provided by authoritative sources, such as mapping agencies (e.g., Swisstopo, Corine Land Cover) and scientific datasets (e.g., the IEEE IADF data fusion contest dataset [177] or the So2Sat LCZ42 dataset [178]).

However, the thematic content of social media posts, i.e., information beyond location and time, contains many subjective estimations, observations, and perceptions. This subjectively colored information on areas or cities offers an approach urban atmospheres, perceptions, or emotions [179]. These applications can be summarized into two categories: human perception of environment, and sentiment analysis. The perception of the environment is more related to vision tasks using social media images, for example mapping animal and plant species [99], and assessing landscape preferences [100] and aesthetics [81], [101]. Urban areas in particular can benefit from this data by collecting citizen sentiments and opinions

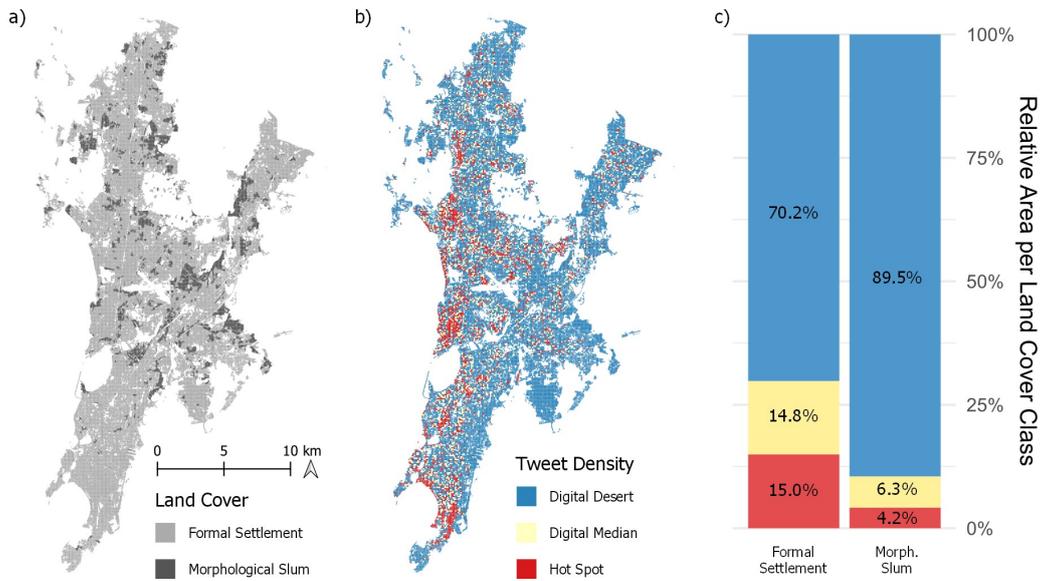

Fig. 8. a) Remotely sensed classification of the urbanized area into formal settlements and non-formal morphologic slums; b) Localization of digital hot spots and deserts based on tweet quantities; c) Proof that digital deserts are more common in morphological slums.

from social media image content and textual descriptions [102], [138].

Sentiment analysis is more related to the processing of text messages. Corpus-linguistic and discourse-analytical methods aim to analyze language in terms of keywords, choice of words, designations of self or external references, discourse topics and subjective evaluations of objects, items, and persons [180]. Investigating people's moods could reveal insights into the problems or challenges of an urban area, potentially important information for city planners or municipal governments. For example, [149] studied the sentiment of Twitter users of urban green spaces in New York City through human annotated tweets. The study revealed that tweets in Manhattan parks exhibit lower sentiment than tweets from the streets. However, in other city districts, the opposite was the case: tweets in parks exhibited more positive sentiments. In addition to that, [150] found that tweets sent from Birmingham, UK, parks mostly expressed happiness and appreciation of nature's beauty. In a recent study, [151] explored the urban Twitter sentiment within the context of the Women's March of 2017 in the US. To examine sentiment, they used the lexicon-based approach VADER, described in subsection IV-B, which specializes in Twitter text.

We demonstrate an example using the term *landscape beauty* in the following paragraphs. When dealing with more "subjective" topics, like the perceived aesthetics of the land, a curated dataset can hardly exist, as *beauty is in the eye of the beholder* and every person has a different perception. Using crowdsourcing is an efficient way to gather data and has been effectively used to gather label points for both objective tasks and subjective opinion tasks (e.g., city perception [152] or, as in our case, landscape beauty [154]). In the case of aesthetics, the advantage of proceeding by crowdsourcing is to develop the ability of capturing subjectivity and learn something about how people feel about nature from the responses.

In the application presented here, which the reader can find in extended form in [155], the ScenicOrNot dataset (SoN, http://scenicornot.datasciencelab.co.uk) is used to make the connection between land use (observed through a series of remote sensing images from the Sentinel-2 satellite) and perceptions of beauty. SoN is a collection of crowdsourced opinions of a series of more than 200,000 images over Great Britain, organized over a 1 km grid. The images are from the Geograph project presented above (https://m.geograph.org.uk); SoN adds perceptive information to these images, by letting volunteers rate each picture with a score between 1 (unsightly) and 10 (beautiful). The ability to predict the scenic quality of an image by using deep learning algorithms was studied recently [124], [181], which then enabled connections to be made between color spaces [154] or objects categories visible in the images [182] and the perception of beauty.

In the application presented here, we aim to study whether land use – an "objective" characteristic visible on a satellite image and for which vast quantities of labeled data exist – influences the perception of beauty. To do so, we re-purpose the model of [182] to predict the scenicness of the landscapes imaged in a Sentinel-2 patch (see Fig. 9). As in [182], the model uses two predictive heads: the first predicting land use as a multi-label problem (i.e., several land use types can be present in the Sentinel-2 patch, but we are not interested in their exact location) and using Corine Land Cover as the ground reference. The outcome of this first predictive head is used by the second prediction task, which estimates the average scenicness for the entire patch. Since the footprint of the Sentinel-2 patch is wider than the precise location of an image, we predict the average scenicness of all the SoN images located within the patch footprint. By using land use as semantic interpretation guidance, we are then able to
(1) tell whether certain types of land use lead to prettier landscapes, and (2) validate the geographical consistency of the scenicness prediction model. The latter is very important since it allows us to question the validity of the interpretable scenicness predictor, for example in cases where the model predicts mountainous land where there are no mountains.

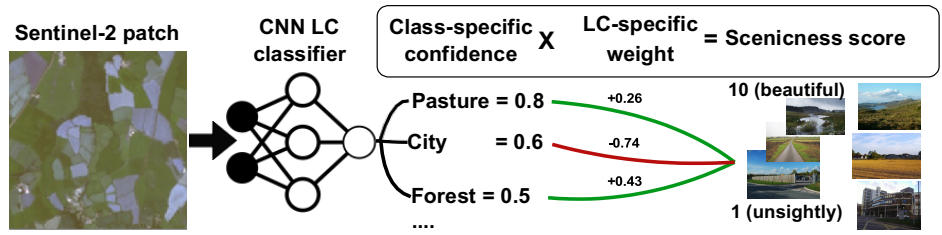

Fig. 9. Intuition behind the interpretable scenicness prediction model in [155]. Photos are obtained from https://www.geograph.org.uk/

The scores provided by volunteers on social media allow us to draw this interpretation map between perceived beauty and land use: in Fig. 10 we display the maximum possible contribution to the final scenicness score. These weights allow the model to find land use-related subtle variations of scenicness around the mean score of the dataset. From the weights one can see that, on average, urban fabrics contribute negatively to landscape aesthetics, while open spaces, non agricultural vegetated areas, and inland wetlands tend to contribute positively.

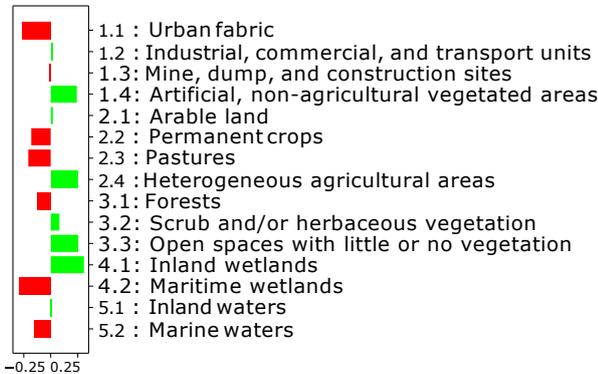

Fig. 10. Weights for each class in the final scenicness prediction layer, as in [155].

Summarizing this application, social media and crowdsourcing tell us something that is generally not available in traditional land prediction tasks: information about the subjective reaction of the users of space. This information is contained in social media data and can be used to learn about perceptions about space and spatial preferences, and potentially can lead to better location- (and customer-) based analytics.

*D. Crisis Response*

During a crisis situation, quickly gaining as much information as possible about the progression of events is of crucial importance. Gathering information is necessary for developing situational awareness, and can mean the difference between life and death. Social media is one source of such information that has started garnering interest in recent years. Twitter users write about disaster preparations, developments, recovery, and many other topics [158]. Retrieving this information can lead to improvements in disaster management strategies. In contrast to most other information sources, social media posts appear nearly immediately whenever there is a new occurrence (as long as telecommunication infrastructure is still intact), and can therefore deliver information very quickly. Such messages can also provide new perspectives that would not be available any other way at this speed, e.g., ground photos. In addition to factual information, social media can offer personal insights into the occurrences, as well as a back-channel to users for relief providers, government agencies, and other official institutions as well as the media. In a 2010 Red Cross study, 69% of Americans said that emergency response agencies should respond to calls for help sent through social media channels [183]. A highly comprehensive overview of social media usage in crisis situations is given in [184].

The difficulty lies in the retrieval and classification of such messages. As described in Section III, the incoming volume of Twitter messages in the live stream is huge. In any given event, the majority of these posts will not be relevant to the event, or useful to service providers. The task of finding social media posts in a crisis may appear clearly defined at first, but quickly becomes more convoluted when attempting an exact definition. Existing publications have defined this problem statement in a variety of ways. As described in [185], research generally focuses on three qualities of messages: whether they relate to the crisis event at all, whether they are relevant, or whether they are informative. All of these questions, but the last two in particular, depend highly on the users of the system, be they affected citizens, family members, the government, news media, or others. In many cases, users are assumed to be relief organizations. In addition, each of these users may be interested in a different use case of the system. For instance, humanitarian and governmental emergency management organizations are interested in understanding "the big picture," whereas local police forces and firefighters desire to find "implicit and explicit requests related to emergency needs that should be fulfilled or serviced as soon as possible" [186]. Moreover, some of these use cases may require a high precision of the detected tweets while possibly missing some important information; others may be more accepting of false alarms while focusing on a high recall.

Several detection approaches have been presented in the literature so far, falling into three categories: filtering by characteristics, crowdsourcing, and machine learning-based. The most obvious strategy is the filtering of tweets by various surface characteristics as shown, for example, in [187]. Keywords and hashtags are used most frequently for this task and often serve as a useful pre-filter. Olteanu et al. developed a lexicon called *CrisisLex* for this purpose [159]. However, this approach easily misses tweets that do not mention the keywords specified in advance, particularly when changes occur or the attention focus shifts during the event, or it may retrieve unrelated data that contains the same keywords [188]. Geo-location is another frequently employed feature that can be useful for retrieving tweets from an area affected by a disaster. However, this approach misses important information that could be coming from a source outside the area, such as help providers or news sources. Additionally, as described in Section II, only a small fraction of tweets is geo-tagged at all, leading to a large quantity of missed tweets from the area [72].

To resolve these problems, crowdsourcing strategies were developed, i.e., asking human volunteers to manually label the data. Established communities of such volunteers can be activated quickly in a disaster event, e.g., the *Standby Task Force* (https://www.standbytaskforce.org/). To facilitate their work, platforms have been developed over the years. One of the most well-known systems is *Ushahidi* (https://www.ushahidi.com). This platform allows people to share situational information in various media, e.g., by text message, by e-mail, and of course by Twitter. Messages can then be tagged with categories relevant to the event. Other examples include *AIDR* [188], which contains automatic analysis tools, or *CrisisTracker* [168]. In *CrisisTracker*, tweets are also collected in real time and clustered by topics so that volunteers can analyze them jointly.

In recent years, approaches based on deep learning techniques have come to the forefront of research. Caragea et al. first employed Convolutional Neural Networks (CNN) for the classification of tweets into those related to flood events and those unrelated [160]. In many of the following approaches, a type of CNN developed by Kim for text classification is used [69], such as in [189]. This method achieves an accuracy of 80% for the classification into related and unrelated tweets. The authors of [161] demonstrate how to implement a system that is able to determine whether a tweet belongs to a class (i.e. crisis event) implicitly defined by a small selection of example tweets by employing few-shot models. The approach is expanded upon in [162].

Once tweets related to a disaster event have been discovered, further analysis steps are possible. A popular next step is the classification into semantic or information type classes, e.g., affected people seeking various types of assistance, media reports, warnings and advice etc. (e.g., [190], [191]). In [163], tweets are clustered by topic at time of publication; these topics develop in real time and can shift. Another way of further discerning between tweets is to distinguish between levels of informativeness or priority (e.g., [167]). It is also possible to perform a more finegrained analysis of particularly interesting properties, such as the sentiment analysis performed on continuous value and time scales during the COVID-19 pandemic in [62] (see also Fig. 11). Other research focuses on the detection of specific events, or types of events (e.g., floods, wildfires, or man-made disasters, e.g., [192]). This can often be helpful when social media is used as an alert system. Apart from these text-based tasks, image analysis can also be a helpful source of information, e.g., for determining the degree of destruction in the aftermath of a disaster [193]. Several datasets have been published for disaster applications to gain insights about the content of social media and to develop detection methods. These include, for example, *CrisisLexT26* [159], [164], *CrisisNLP* [165], *CrisisMMD* [166] (also including images), and *TREC-IS 2019A* [167]. The annual *TREC Incident Streams* challenge invites researchers to submit novel methods for classifying disaster-related tweets, and continuously provides new labeled data (http://dcs.gla.ac.uk/richardm/TREC_IS/). A more detailed overview on the detection of Twitter messages in crisis events is given in [194].

VIII. ETHICS OF RESEARCH WITH SOCIAL MEDIA DATA

In the world of (business) ethics, the observer effect may translate to what is known as ethical relativism. Under the doctrine of relativism (and also in practical reality), ethical issues – questions of right or wrong, good or bad – vary based on who is observing and the context within which the observation is being made [195]. Ethical issues in the context of social media data mining have already been flagged in diverse contexts, notably in the context of medical data. The primary aim of this segment is to broadly identify which of the ethical concerns that have been raised in the general context of social media data mining are likely to apply, in the present or in the future, to geo-information harvesting from social media.[4]

*A. Ethical Issues Commonly Flagged in Social Media Data Harvesting*

Significant research has been undertaken to identify ethical issues that (may) arise from data mining per se, as well as from the labeling, analysis, and use of the data (e.g., in AI or machine learning models). The most commonly identified issues include:

a) Privacy
b) Stigmatization
c) Data veracity
d) Bias (including bias in training data)
e) Transparency and explainability

---

[4]The Ethics Guidelines for Trustworthy AI, which provides a "concrete and non-exhaustive Trustworthy AI assessment list", also recognizes that the list will "need to be tailored to the specific use case of the AI system." The Guidelines also suggest that "in addition to this horizontal framework, a sectoral approach is needed, given the context-specificity of AI systems," and this needs exploration. This article does not address ethical issues in AI. Nevertheless, as the geo-information data mined from social media platforms is largely used to train AI/ML models, this observation made by the Ethics Guidelines for Trustworthy AI is relevant to the overview given here. References from these guidelines (referred to hereinafter as "Trustworthy AI Guidelines") have been made, if relevant, in footnotes.

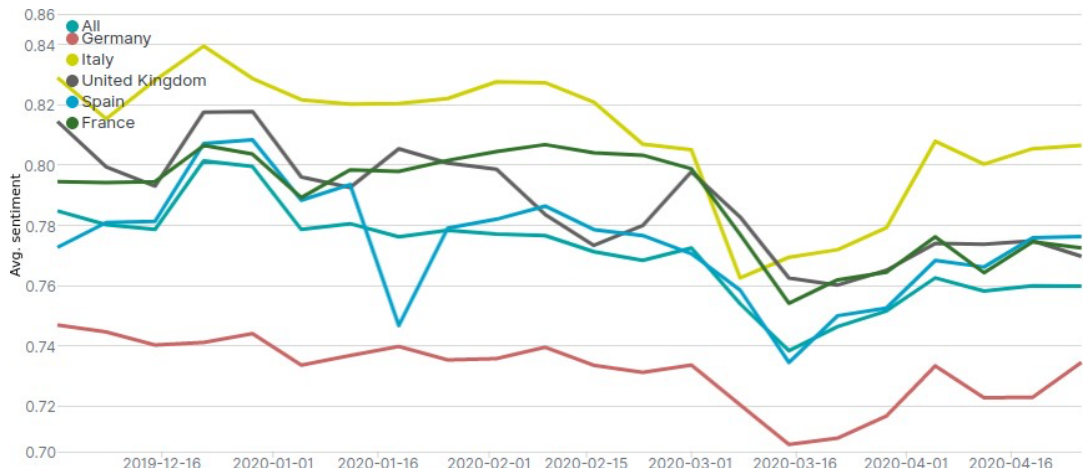

Fig. 11. Sentiment development of tweets in various European countries during the first months of the COVID-19 pandemic [62].

f) Security (including in the context of data storage)
g) Accountability and democratic creation of standards

In addition to the above issues, in the context of geo-information data harvesting for use in EO/RS applications, we may also need to consider the issues of national security and national sovereignty. These issues, however, may or may not arise depending on the use case (end result sought) and on how other issues such as security and data dissemination are handled.

The seven issues listed above have been raised in very diverse contexts (e.g., medical data mining from social media) in existing literature. In the following sub-section, we look at each of these briefly in the context of geo-information data harvested from social media, especially where such data is intended for use in EO or RS applications.

### B. Ethical Issues in Geo-information Data Harvesting

The most important ethical issues that may arise in the context of geo-information data mining for use in EO or RS applications are discussed below. However, researchers must be mindful that this is not a comprehensive list and new issues are likely to arise as technology evolves and new use cases emerge.

*a) Privacy:* Including and beyond the scope of the EU General Data Protection Regulation (GDPR), privacy of persons whose data is mined from various social media platforms needs to be protected at all costs and its misuse prevented. The argument that the data is already public (e.g., as it was posted on Twitter) may not be adequate to avoid or overcome ethical concerns of privacy [196], [197]. Yet, the nature of geo-information data mining for use in EO/RS applications is often very different from data mining for other purposes: for example, geo-information data mined with the aim of labeling buildings or identifying areas where rescue efforts (see Section VII) need to be expedited. Indeed, in emergency situations, tweeting individuals may actively want their identity and location to be accurately discoverable so that help can reach them or their community. Yet, to the extent that comments associated with texts and images (even if these relate to buildings or geographic regions) can facilitate social, economic, political, or racial profiling of individuals, precautionary measures such as anonymization and non-reproduction of verbatim texts are essential. This is especially true of any applications (such as building labeling) that do not require the identity e.g. of the individual tweeter to be known. Further, cultural sensitivities need to be taken into account while collecting data – some cultures may be less tolerant than others to the very act of collecting personal (geographic location) information of individual social media users. These cultural sensitivities might exist beyond the scope of local/national laws.

*b) Stigmatization:* Closely linked to the issue of privacy is the issue of stigmatization [198]. Although most often raised in the context of medical data mining, in the context of geo-information data harvesting, ethical concerns may arise at the more advanced stage, i.e., when the resulting data is used to create population density maps or for labeling areas that are currently not labeled in maps, such as slum areas. Here, labeling (or mislabeling) of social media data such that specific areas are labeled as slums can lead to stigmatization, not (only) at individual, but at community or national levels. The issues of stigmatization also links up to issues of bias and transparency, as well as data veracity, as discussed in the following points. From the perspective of ethical opportunities, on the other hand, responsible and constructive labelling can support development of economically weaker (rural and urban) regions, or expansion of urban green areas by directing government funding to those regions.

*c) Data Veracity:* Although data volume, variety and velocity are relevant for data management (compare Section III), data veracity [199], [200] is equally relevant for accurate and reliable end results. Undoubtedly, the nature of big data, and appropriate programming of AI/ML models that use this data, may minimize the harm caused by partially inaccurate data. However, as the volume of inaccurate or false data increases, the reliability of the end result decreases. The problem of data veracity in social media data mining may not arise from deliberate dishonest statements or uploads of doctored images by social media users. However, it is likely that a user may tweet about one location while sitting in a completely different location. Similarly, NLP still needs to evolve to be able to fully recognize and differentiate between regular everyday language and satire, sarcasm, jokes, or irony. In politically controversial circumstances (e.g., man-made calamities or controversial emergency situations), the use of these linguistic devises may increase. In the current COVID situation, for example, with some segments of the population turning against governmental safety regulations, protests from people may be visible in diverse forms on social media platforms. Recent governmental bans of weekend protests against COVID measures [201], (which were later overturned by a court ruling [202]), and alleged inaccuracies in reporting of number of participants in previous protests [201] also highlight the need to use geo-information from social media data carefully.

*d) Bias in Training Data:* Published work has also highlighted the ethical issues that can or do arise from using data from any source (including from social media platforms) to train (inter alia) machine learning models [203], [204]. Several recent incidents have highlighted the (negative) unintended impact that biased training data can have on ML based tools. Amazon's AI-based recruitment tool that showed a bias against women, is one example [203]. More recently, the algorithm PULSE that uses a process called "upscaling" to convert low resolution pixelated images to high resolution images, was also found to be biased. When used to enhance facial images of real human beings, the resulting faces were "distinctly white," even when the input low-resolution images were of Barack Obama or Lucy Liu [205].

Right from the start of any research endeavor, bias can result from selection (inclusion or exclusion) of specific data sources. For example, in different countries, different social media platforms are popular and the selection of any one (at the exclusion of others) can lead to bias (e.g., Flickr in the USA; VK (VKontakte) in Russia; Weibo in China [206], [207]).

It is, therefore, necessary for researchers to be aware of and disclose the reasons for selecting specific data sources over others. Often, the main reason may be the access rights given by some platform providers based on their existing end user licenses. Not all social media platforms grant such access. This convenience-based platform selection may itself lead to biases. In Flickr, for example, the type of details (e.g., camera specifications) that can be entered by users suggests that it is used only by a very specialized, artistic audience (see Section V). Accordingly, if determinations of building or region aesthetics are based on Flickr data, there is a high likelihood of bias due to individual "artistic" depictions of specific regions or locations. Similarly, in Twitter, a significant percentage of the text and data shared may specifically relate to places Twitter users are proud to be associated with or found at, e.g., famous buildings, or popular tourist destinations.

In 2020, an estimated 3.6 billion people used various social media platforms [208]. Yet, in the context of geo-information mining from social media data, where one of the aims is to train models to make predictions for regions from where data may not be available, researchers are well aware of the digital divide [209]–[211]. In developing countries, residents of rural and slum areas may increasingly use social media platforms in the future (efforts to create social media platforms for use in local languages are underway). However, only a small minority of the global population uses "popular" social media platforms [209]. The elderly and the disabled may also be severely underrepresented in data sets taken from social media platforms. Platforms that by their nature permit or encourage public sharing of information or images (e.g., Twitter) are used by an even more limited number and "category" of people. When data harvested from such sources is used to train ML models to, for example, identify "slum areas" or estimate building level population density, several precautions must be exercised, and several limitations of the dataset accurately disclosed so as to alert users to potential gaps or inaccuracies in the predictions.

Let us say, for example, that social media text data from Twitter (see Section IV) users is used to train ML models and these models are used to predict building level population density in various regions of the world. Twitter demographics reveals that 63% of Twitter users are above 35 years of age [211]. Will models trained on Twitter data then be able to accurately predict building level population density in countries like India where an estimated 65% of the population is below 35 years of age [212]?

Following (platform) selection bias, bias can also creep into the training data at the time of labeling – known or unknown personal concepts, prejudices, and cultural beliefs of the "observer," i.e., the one placing the labels, need to be identified and comprehensively declared at the outset [209], [213].

*e) Transparency and Explainability:* Other issues that are frequently flagged, particularly in the context of AI/ML applications that utilize big data, relate to ensuring fairness, (decisional) transparency [204], and explainability [214]. Given issues such as stigmatization (discussed above), and security and sovereignty (see below) that may arise from labelling of data, it is important that AI algorithms are able to trace back their steps from the end result they deliver to the specific steps taken to reach that result. The extent to which this is necessary and possible in the context of various EO/RS applications utilizing geo-information data from social media, needs to be examined.

*f) Security (including in the context of data storage):* It is necessary to distinguish between the immediate use to which the harvested social media data is put, and other uses which the data, once stored, may be put to in the future. To the extent the data collected from social media is used, for example, for semantic labelling of buildings, or labels that link to the beauty and qualities of a specific land/area (see Section VII), or for providing live and immediate assistance during a crisis, legal and ethical concerns (e.g., linked to the topic of privacy or the right to be forgotten) may not arise. In fact, particularly in crisis situations, the ethical view would be that concerns about privacy (or other peripheral issues) should not cause rescue or support operations to be prevented, delayed, or discontinued. However, if the same data is stored, and later shared publicly for other purposes (e.g., profiling voters), ethical as well as legal issues (e.g., under the GDPR) may arise. Accordingly, while public and open sharing/dissemination of stored data may appear to be a good policy, the consequences of such sharing need to be studied closely to avoid any downstream ethical issues from arising. Further, storage of personal data is regulated under the GDPR. Beyond legal regulations, fairness requires that personal data of individual users not be stored (or shared) beyond necessary periods of time, and beyond the (legitimate) reasons for which it was originally gathered. This is particularly important also when the original tweet, video, image, or other social media data is deleted by its creator. Data must also be stored in a way that does not permit the identification of the individual concerned (unless the consent of the individual is legitimately obtained and is necessary given the objective of the application).

Security also requires that data be stored in a way that prevents unauthorized access. Those authorized to access stored data must also be bound by legal obligations and ethical guidelines.

*g) Accountability and Democratic Creation of Standards:* At the (higher) level of governance, the ethical issues that have been identified include content moderation, platform regulation, and creation of standards (e.g., in the context of Facebook's photo-matching algorithm) [203]. Indeed, applications using big data to guide governmental decision making, city planning, and the like need to ensure that those creating standards or labelling guides are accountable and made ethically responsible for any resulting inequities. Standards and labeling guides can be created after taking multiple stakeholders' views into account. They must also abide by the fundamental principles of equity, fairness, honesty, and integrity.

*h) National Security and National Sovereignty:* In the context of geo-information harvesting for enhancing EO or RS data, issues of personal security are perhaps less relevant than issues of national or regional security. Indeed, project-specific, institutional, national, or international regulations on maximum image resolution can be used to ensure that identifying individuals, including facial recognition or identification of car plates, is not permitted. Further, where remote sensing data is fused with social media data to predict or gather information (rapidly) about natural or man-made calamities in a region, the ethical opportunities provided by such geo-information data harvesting are immense.

However, when such fused data is used to estimate building level population density or for semantic labelling of building types, it is possible that wide-spread dissemination of such information could raise issues of (national) security. For example, crowded areas or commercial buildings could be easily identified by terrorist groups to plan attacks or threats to disrupt civil life or cause maximum damage.

Similarly, EO/RS and social media data can be used to pressurize national governments to prioritize policies that seem important from an international or standardized standpoint, but may not be urgent or important in local/national contexts. This can compromise national sovereignty and a people's right to self-determination.

Ethical issues linked to EO/RS research and development are evolving and coming into the limelight rather slowly. It is important that research institutions as well as private enterprises working with EO/RS, especially, but not exclusively, in combination with geotagged social media data, educate themselves on emerging ethical issues and remind mindful of ethical issues that can arise in this rapidly growing field of research.

## IX. Conclusion and future trends

There can be no doubt that the massive volume of data from social media is a gold mine of geo-information. In particular, thanks to its complementarity with remote sensing data, fusion offers new perspectives for a number of geographic applications. In this article we have discussed crucial aspects of geo-information harvesting from social media data, ranging from data availability, data management, geo-information retrieval algorithms, and its fusion with remote sensing data, while showcasing geographic use cases and raising ethical considerations.

Looking forward, several major challenges remain unsolved:

- *Unstructured data:* For decades, remote sensing has largely focused on information retrieval from satellite images rasterized in geographical space or 3D point clouds, data that is unstructured yet still bound to the geometric shape of Earth surfaces that can be easily modeled as dense volumetric grids [215]. Unlike such Euclidean data, social media data exhibits a more complex structure. Billions of people interact daily and leave digital traces. While analyzing these data at global scale is very promising, numerous algorithmic and methodological challenges remain to index such non-Euclidean data and link them with other geo-data, via either their geo-tags or their semantic contents.
- *No geo-tag, inaccurate geo-tag, wrong geo-tag:* Each pixel in remote sensing data corresponds to an accurate spatial coordinate, which facilitates the fusion of pixel information with other sensors or other sources of data. However, geolocation information on social media data is still very limited. Only about 0.87% to 3% of the data is geo-tagged [216], and even that exhibits very diverse geolocation accuracy. Taking Flickr as an example, in [217], it is reported that the average distance between the gold standard location and the provided location is 11–13 meters for popular venues, and approximately 47–167 meters for less popular venues. In addition, image or text posting at a certain geolocation may not directly reflect the activities or ground-level information at the time and location of the posting: for example, an image of the Eiffel Tower can be posted from any other location.
- *Information mining:* Social media users upload very diverse content. Any geographic application requires mining of relevant data. This is the foundation of any further analysis. Only then can inference, spatial and thematic classification, and change detection be performed. While many automatic filtering approaches focus on a classification task, this is often not sufficient for real-world use cases such as disasters and crises. Here, it is frequently more important to detect emerging topics or tweets with a news value over known situational information. This also means that taking this known context into account is vital to the filtering approach. An implicit approach for detecting relevant tweets based on few-shot learning is shown in [161], while first approaches for detecting novelty and emerging topics in the tweet stream can be found in [163], [218], [219]. An overview of the shift towards adaptable approaches to detect so-called "actionable" tweets is given in [194].

- *Uncontrolled quality:* Almost every aspect of social media data allows manual editing by their authors: time, location, and content. For example, tweets can be scheduled for publication at a certain date and time or photos are tagged manually because the camera has no GPS sensor. Moreover, if users enable automatic geo-tagging, their posts can be unrelated to their location.

  In the end, data quality is highly heterogeneous in social media datasets: some posts have a direct relation to their surroundings, while others have no connection at all or one that is only visible for a group of users. Depending on the scientific question different filtering approaches have to be applied, e.g., language, content, or time and date.

  Beyond filtering, uncertainty measures in the models can help to identify areas where the spatial prediction has little support from the data or can even create more accurate spatial knowledge by taking the inherent noise as a feature.

- *Adversarial attacks:* A particularly aggravating consequence of the lack of control over data quality may be the risk of adversarial attacks against automatic systems for the analysis of big social media data sources. As discussed in [220], [221], this topic has started to come to the forefront of NLP research after beginning in computervision. In short, a malicious attacker may manipulate the incoming data in a way that is not perceptible to humans, but may lead to false results. This is much easier to do when the attacker has access to the system (e.g. neural network model), but can even be attempted in a black-box setting. Obviously risky tasks here include the geo-related usage of social media data in the context of political issues, such as elections or protests, or during man-made conflicts. As a starting point, awareness of this possibility as well as close monitoring of model behavior are important. Obscuring the models may help, but runs counter to general desires for the transparency of AI systems. Ultimately, models robust to these manipulations are necessary and will be a focus of future research.

- *Opportunistic data:* The distribution of social media imagery and texts are opportunistic and geographically very non-uniformly distributed, e.g., tourist attractions such as the Eiffel Tower can be found much more frequently than images of a particular slum. With respect to geographic regions, as of January 2021 the global social network penetration rate (that is, active users versus the total population) by region reveals a global average of 53.6%, with East Asia and North America having the highest penetration rate at 71 and 69% respectively, and with Middle Africa having the lowest penetration rate of only 8% [222].

- *Ethical considerations:* In relation to ethical issues that already have arisen, or may arise in the near future, in the context of geo-information harvesting from social media data, it is necessary to bear in mind that a significant amount of work being currently undertaken in this sphere is exploratory. Accordingly, several ethical issues may yet be moot or unknown. Accordingly, we recommend the development of a more comprehensive approach and framework that can help categorize and flag ethical issues that (may) arise in the future at various stages of research, development, and innovation with geo-information data from social media.

To move community remote sensing with social media forward, joint forces are crucial. Thus, we would like to emphasize the importance of "open science – open data".

ACKNOWLEDGMENT

This work is jointly supported by the European Research Council (ERC) under the European Union's Horizon 2020 research and innovation programme (grant agreement No. [ERC-2016-StG-714087], Acronym: *So2Sat*), by the Helmholtz Association through the Framework of Helmholtz Excellent Professorship "Data Science in Earth Observation - Big Data Fusion for Urban Research" (grant number: W2-W3-100), and by the German Federal Ministry of Education and Research (BMBF) in the framework of the international future AI lab "AI4EO – Artificial Intelligence for Earth Observation: Reasoning, Uncertainties, Ethics and Beyond" (Grant number: 01DD20001). The authors would like to thank Prof. Christoph Lütge and Caitlin Corrigan for their comments on the first draft of the ethics segment, and Ms. Julia Köninger for excellent research assistance for this segment.